\documentclass[letterpaper, 10 pt, conference]{ieeeconf}  

\IEEEoverridecommandlockouts                              
\usepackage{subcaption}
\usepackage{mwe}
\usepackage{graphics} 
\usepackage{bm}
\usepackage{siunitx}
\usepackage{relsize}
\usepackage{amsmath} 
\usepackage{amssymb}  
\usepackage[ruled,vlined]{algorithm2e} 
\usepackage{booktabs}
\usepackage{siunitx}
\usepackage{multirow}
\usepackage{makecell}
\usepackage{subcaption} 
\usepackage{hyperref}
\usepackage[usenames, dvipsnames,table,xcdraw]{xcolor}
\usepackage{pifont}

\usepackage{tabularx}
\UseRawInputEncoding

\usepackage{tabularx}

\usepackage[capitalise]{cleveref}
\usepackage{bm}
\usepackage{mathtools}

\title{\LARGE \bf  Unlocking the Potential of Soft Actor-Critic for Imitation Learning}
\author{Nayari Marie Lessa$^{1,2*}$, Melya Boukheddimi$^{1}$, and Frank Kirchner$^{1,2}$
\thanks{{$^{*}$Corresponding author}: \href{mailto:lessa@uni-bremen.de}{lessa@uni-bremen.de}}
\thanks{$^{1}$Robotics Innovation Center, DFKI GmbH, 28359 Bremen, Germany.}
\thanks{$^{2}$University of Bremen, Robotics Research Group, 28359 Bremen, Germany.}
}

\begin{document}
\newcommand{\robotName}{RH5V2}

\newcommand{\mvec}[1]{\bm{#1}}
\newcommand{\vc}[1]{\mathbf{\mathbf{#1}}}

\newcommand{\q}{\textbf{q}}
\newcommand{\dq}{\dot{\q}}
\newcommand{\ddq}{\ddot{\q}}


\newcommand{\Mass}{\mathbf{M}}
\newcommand{\Bias}{\mathbf{b}}
\newcommand{\Gravity}{\mathbf{g}}
\newcommand{\Force}{\mathbf{\lambda}}
\newcommand{\Torque}{\mathbf{\tau}}
\newcommand{\Jac}{\mathbf{J}}

\newcommand{\BIN}{\begin{bmatrix}}
\newcommand{\BOUT}{\end{bmatrix}}

\newcommand{\sref}[1]{Sec~\ref{#1}}
\newcommand{\eref}[1]{(\ref{#1})}
\newcommand{\fref}[1]{Fig.~\ref{#1}}
\newcommand{\tref}[1]{Table~\ref{#1}}
\newcommand{\equationref}[1]{Equ.~\ref{#1}}
\newcommand{\state}{\mathbf{x}}
\newcommand{\ctrl}{\mathbf{u}}
\newcommand{\dynsys}{\mathbf{f}}

\newcommand{\qTr}{\underline{\q}}
\newcommand{\dqTr}{\underline{\dq}}
\newcommand{\ddqTr}{\underline{\ddq}}
\newcommand{\TorqueTr}{\underline{\Torque}}

\newcommand{\costl}{l}
\newcommand{\dts}{\Delta t_s}
\newcommand{\st}{\text{subject to}}

\maketitle
\thispagestyle{empty}
\pagestyle{empty}
\begin{abstract}
Learning-based methods have enabled robots to acquire bio-inspired movements with increasing levels of naturalness and adaptability. Among these, Imitation Learning (IL) has proven effective in transferring complex motion patterns from animals to robotic systems. However, current state-of-the-art frameworks predominantly rely on Proximal Policy Optimization (PPO), an on-policy algorithm that prioritizes stability over sample efficiency and policy generalization. This paper proposes a novel IL framework that combines Adversarial Motion Priors (AMP) with the off-policy Soft Actor-Critic (SAC) algorithm to overcome these limitations.
This integration leverages replay-driven learning and entropy-regularized exploration, enabling naturalistic behavior and task execution, improving data efficiency and robustness.
We evaluate the proposed approach (AMP+SAC) on quadruped gaits involving multiple reference motions and diverse terrains. Experimental results demonstrate that the proposed framework not only maintains stable task execution but also achieves higher imitation rewards compared to the widely used AMP+PPO method. These findings highlight the potential of an off-policy IL formulation for advancing motion generation in robotics.
\end{abstract}
\section{Introduction} 
\label{sec:intro}
The past decade has witnessed remarkable progress in learning-based algorithms for robotics, with applications spanning a wide range of domains. These advances aim to endow robots with greater intelligence and adaptability, enabling them to perform tasks with smooth, natural, and human or animal--like motions \cite{10415857, qin2025integrating, eren2024tiny}. Among the many research directions, bio-inspired robotics has emerged as particularly impactful, influencing both mechanical design and motion generation \cite{zhang2024imitation}. Within this context, IL plays a pivotal role in generating bio-inspired behaviors, enabling robots to reproduce complex motion patterns during task execution \cite{budiarto2024design}.
IL has been successfully applied across diverse areas of robotics. For example, in medical applications robotic arms replicate the precise motions of expert physicians \cite{10884876}. While in animal-inspired locomotion, quadruped robots have learned gaits from natural counterparts \cite{10752370}. A variety of IL approaches have been explored. For instance, natural motion generation is proposed by \cite{li2024ai} an approach that learns a CPG-based controller for humanoid gait generation. Similarly, model-based strategies have been developed. \cite{reske2021imitation} Integrated Model Predictive Control (MPC) with Multi-Layer Perceptron (MLP) experts to generate diverse quadruped gaits, while \cite{10268037} combined MPC with deep reinforcement learning (RL) to improve locomotion. 
\begin{figure}[h]
	\centering
	\includegraphics[width =0.45\textwidth,height=5cm]{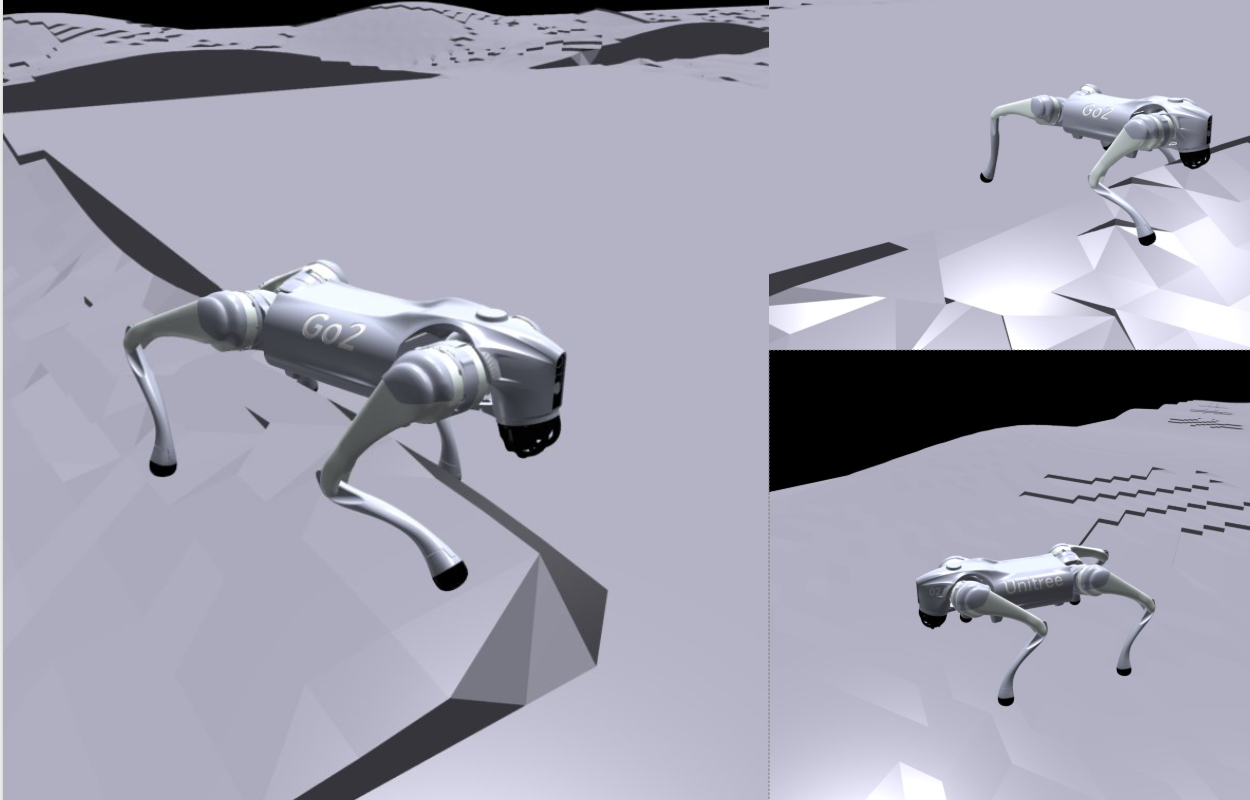}
\caption{Snapshots of an obtained AMP+SAC imitation learning policy performing multiple gaits across different terrains.}
	\label{fig:chinUp}
\end{figure}  
Although effective, these methods mainly rely on optimal trajectory tracking and often neglect the dynamic variability inherent in animal-inspired motion.
To address these limitations, recent research has combined IL with RL, often leveraging real animal motion dataset as references. For example, \cite{chen2025between} proposed "in-between" an IL approach that generated dynamically consistent gaits with precise velocity tracking on quadruped robots. Similarly, \cite{zhang2024learning} introduced an RL-based IL framework trained on dog motion data, where a trajectory encoder–decoder was coupled with PPO \cite{schulman2017proximal}, enabling the reproduction of highly dynamic behaviors on real hardware.
Among RL algorithms, PPO has become the dominant on-policy method in IL, primarily due to its stability and efficiency \cite{luo2024behavior}. While stable, PPO is limited in sample efficiency and policy generalization. A wide range of PPO-based IL frameworks have been proposed to transfer animal-inspired behaviors to quadruped robots. Early work demonstrated the feasibility of this paradigm, such as \cite{peng2020learning}, which trained dog-inspired policies and successfully deployed them on physical systems.
Building on this, \cite{li2022versatile} introduced CASSI, an unsupervised adversarial imitation framework coupled with PPO that enabled quadrupeds to acquire a diverse repertoire of behaviors. 
To further enhance robustness, \cite{li2023learning} augmented PPO with a reference motion encoder, improving locomotion on challenging terrains. \cite{han2024lifelike} proposed a hierarchical RL framework in which animal motions were represented as discrete latent embeddings, enabling successful deployment of highly dynamic behaviors on quadruped robots. More recently, \cite{xiao2025stable} leveraged domain randomization during PPO training to obtain robust policies capable of adapting to diverse motions and terrains.
Despite this remarkable progress, most efforts remain focused on performance optimization and sim-to-real transfer, with comparatively less attention given to exploring alternative algorithmic formulations and design choices within IL frameworks. In this work, we aim to address this gap. Specifically, we propose a framework that replaces the commonly used on-policy PPO with the off-policy SAC algorithm \cite{haarnoja2018soft}, while combining it with AMP \cite{peng2021amp, escontrela2022adversarial} to provide structured imitation guidance.
Unlike PPO, SAC benefits from its off-policy nature, allowing more efficient exploration and broader policy generalization. These properties are particularly important for robots that must operate in unconstrained environments, adapting to diverse motions and terrains while maintaining smooth, natural behaviors. As a use case, we evaluate quadruped locomotion across multiple motions and terrains. Including a performance benchmarking  against a baseline AMP+PPO implementation.
\paragraph*{Contributions}
The main contributions of this paper are:
\begin{itemize}
\item A novel IL framework that integrates AMP with the off-policy SAC algorithm, enabling robust task execution while preserving natural, animal-inspired motion.
\item Extensive evaluations on quadruped locomotion tasks involving  multi-motions and varying terrains, demonstrating the framework’s ability to generalize beyond reference trajectories.
\item Performance comparison against a widely used AMP+PPO baseline, showing that our approach achieves superior imitation rewards.
\end{itemize}
\paragraph*{Organization}
The remainder of this paper is organized as follows.
Section~\ref{sec:methodology} introduces the proposed methodology.
Section~\ref{sec:implementation} describes the experimentation setup.
Section~\ref{sec:results} presents the results and comparisons with the baseline.
Finally, Section~\ref{sec:conclusion} summarizes the findings and limitations and outlines future research directions.
\section{Methodology} 
\label{sec:methodology}
\subsection{Synopsis}
\begin{figure}[h]
	\centering
	\includegraphics[width=0.45\textwidth]{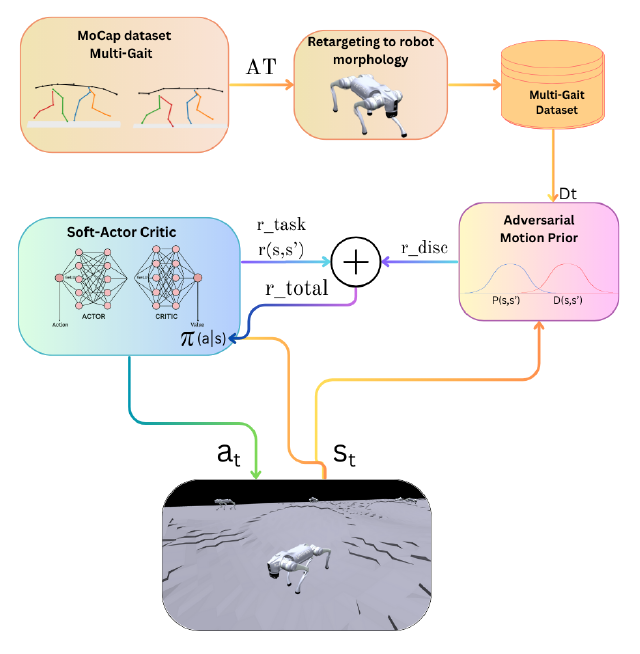}  
	\caption{Pipeline of the AMP+SAC imitation learning.}
	\label{fig:pipeline}
\end{figure}

In this work, we aim to train a robot to imitate animal behaviors using a combination of reinforcement learning and adversarial imitation. 
The overview of the proposed methodology is illustrated in \fref{fig:pipeline}. The pipeline is categorized in two sequential stages: (1) motion capture processing for extracting reference kinematic features, and 
(2) adversarial imitation learning. The latter stage is itself composed of two components: training a discriminator to differentiate expert from agent motions, and optimizing a policy via SAC. 
\subsection{Reference Data Processing}
In order to perform our training, a data processing step is necessary, for that reason, the raw dataset is
 pre-processed using affine transformations to account for scale differences and foot offsets. To ensure kinematic feasibility, we converted the 3D joint positions into joint angles via inverse kinematics and computed joint
 velocities as well as base linear and angular velocities. This procedure yields a sequence of joint angle vectors ${q_t}_{t=1}^{T}$, where $T$ is the number of frames in the motion clip. The final reference motion dataset $\mathcal{D}$ consists of root position and orientation, joint angles, joint velocities, and foot positions, all expressed in the local frame of the robot.
\subsection{Discriminator-Guided Imitation with Soft Actor Critic}
We formulate the locomotion imitation learning problem as a Markov Decision Process (MDP) defined by a tuple $
\mathcal{M} = (\mathcal{S}, \mathcal{A}, P, r, \gamma)$, where $\mathcal{S}$ is the state space, $\mathcal{A}$ is the action space, $P(s'|s,a)$ denotes the transition dynamics, $r(s,a, s')$ is the reward function, and $\gamma \in [0,1]$ is the discount factor. 
The goal of Reinforcement learning is to find a policy $\pi_\theta(a|s)$ parameterized by $\theta$ that maximizes the expected cumulative discounted reward
 $J(\pi) = \mathbb{E}_{\pi}[\sum_{t=0}^{T} \gamma^t r(s_t,a_t)]$ over a finite horizon $T$.
To solve the locomotion control problem, we adopt the SAC algorithm~\cite{haarnoja2018soft}, an off-policy actor-critic method that augments the reward maximization objective with an entropy regularization term. This encourages exploration and improves robustness of the learned policy.
The SAC objective is defined as:
\begin{equation}
\begin{aligned}
    J(\pi) = \sum_{t=0}^{T} \mathbb{E}_{(s_t, a_t)\sim \pi}\big[ r(s_t,a_t) + \alpha \, \mathcal{H}(\pi(\cdot|s_t)) \big],
\end{aligned}
\end{equation}
where $\mathcal{H}(\pi(\cdot|s))$ denotes the entropy of the policy $\pi_\theta$, and $\alpha$ is the temperature coefficient that controls the trade-off between maximizing the expected return and encouraging exploration. 
In our framework, the actor $\pi_\theta(a|s)$ is parameterized by a multi-layer perceptron (MLP). The network receives normalized states at time $t$ and outputs the mean and log-standard deviation of a Gaussian distribution. An action is then sampled from this distribution and passed through a $\tanh$ squashing function to ensure outputs lie within valid bounds, a process implemented using the re-parameterization trick to enable end-to-end gradient back-propagation. The actor outputs continuous joint commands for the robot. 
The critic consists of two independently initialized Q-networks, $Q_{\psi_1}(s,a)$ and $Q_{\psi_2}(s,a)$, each implemented as an MLP. These are trained to estimate the expected soft Q-value, which includes the expected return plus an entropy bonus, for state-action pairs.
A discriminator network $\mathcal{D}_{\phi}$ is trained to distinguish between state transitions $(s, s')$ generated by the policy $\pi$ and those from the expert dataset $\mathcal{D}$. 
We employ Adversarial Motion Priors (AMP) \cite{peng2021amp} to train the discriminator objective:
\begin{equation}
\begin{aligned}
\arg \min_{\phi} \; & 
\mathbb{E}_{(s, s') \sim \mathcal{D}} \left[ \left(D_{\phi}(s, s') - 1 \right)^{2} \right] \\
&+ \mathbb{E}_{(s, s') \sim \pi_\theta(s,a)} \left[ \left(D_{\phi}(s, s') + 1 \right)^{2} \right] \\
&+ \frac{w^{\mathrm{gp}}}{2} \; \mathbb{E}_{(s, s') \sim \mathcal{D}} 
\left[ \left\| \nabla_{\phi} D_{\phi}(s, s') \right\|^{2} \right],
\end{aligned}
\label{equ:disc}
\end{equation}
where the first two terms form a least-squares adversarial loss objective to differ the states transitions, and the last term is the gradient penalty regularization controlled by 
the weight $w^{\mathrm{gp}}$ to prevent over-fitting to the reference dataset, improving stability during adversarial training.
\subsubsection{Actor objective with AMP shaping}
In SAC, the actor objective  $\mathcal{J}_{\pi}(\theta)$ optimizes the policy $\pi_\theta$ to maximize expected future return while simultaneously enforcing entropy maximization. Formally, the actor is trained by minimizing the Kullback-Leibler divergence between the entropy-augmented reward, 
weighted by temperature $\alpha$, and the state-action value estimate $Q_{\psi}(s,a)$:
\begin{equation}
\mathcal{J}_{\pi}(\theta)
= \mathbb{E}_{s\sim \mathcal{D}}\ \big[\mathbb{E}_{a\sim \pi_\theta(\cdot|s)}
\big[\alpha \log \pi_\theta(a|s) - Q_{\psi}(s,a)\big]\big].
\end{equation}
To integrate AMP with SAC the actor loss function is augmented with adversarial objectives:
\begin{equation}
\mathcal{J}{\pi}^{\text{AMP}}(\theta) = \mathcal{J}{\pi}(\theta) + \lambda_{\text{AMP}} \cdot \mathcal{L}_{\text{AMP}} + \lambda_{\text{grad}} \cdot \mathcal{L}_{\text{grad}},
\end{equation}
where $\mathcal{L}_{\text{AMP}}$​ is the adversarial loss from the AMP discriminator, encouraging policy-generated state transitions to resemble expert demonstrations, 
and $\mathcal{L}_{\text{grad}}$​ is the gradient penalty loss that ensures training stability for the discriminator. The coefficients $\lambda_{\text{AMP}}$​ and $\mathcal{L}_{\text{grad}}$​ control the weighting of these additional terms.
This formulation enables the policy to acquire complex skills through imitation while preserving the robustness and exploration guarantees of maximum entropy RL.
\subsection{Task and Imitation Reward Formulation}
The adversarial framework minimizes reward engineering by replacing explicit trajectory tracking with a learned reward signal. 
A discriminator $D_\phi$​ provides this signal, guiding the policy to imitate the expert by rewarding it for generating states that 
$D_\phi$​ classifies as real. This AMP reward is defined as:\begin{equation}
\label{equ:value}
r(s_t,s'_t) = \mathcal{W}_{\text{AMP}}\,\max\Big[\,1 - 0.25(D_\phi([s,s']) - 1)^2 \,\Big], 
\end{equation}
where $\mathcal{W}_{\text{AMP}}$ is a scalar coefficient that balances the importance of imitation against the task reward.
The transformation $(D_\phi(\cdot) - 1)^2$ and the $\max$ operation serve to bound the reward within the range [0,1] to stable learning.
Our goal is to evaluate the locomotion imitation while promoting a task-oriented and physically feasible behaviors on the robot. Therefore, the adversarial reward is combined with a separate locomotion reward $r_{\text{task}}(s, a, s')$, 
composing by a forward command velocity $\vec{v}_{\text{t}} = [\vec{v}^{x}_{\text{t}}]$ specified in the base frame, and additional terms that penalize undesired motions such as excessive lateral drift, high base angular velocity around the x,y,z-axis, and position limit violations.
The total reward $r_t$ at each time-step is therefore a weighted sum of these two components:
\begin{equation}
r(s_t, a_t) 
= \mathcal{W}\, r_{\text{task}}(s,a,s'), 
+ \mathcal{W}_{\text{AMP}}\, r_(s_t, s'_t),
\label{eq:reward-total}
\end{equation}
\begin{figure*}[h]
	\centering
	\includegraphics[width =0.99\textwidth]{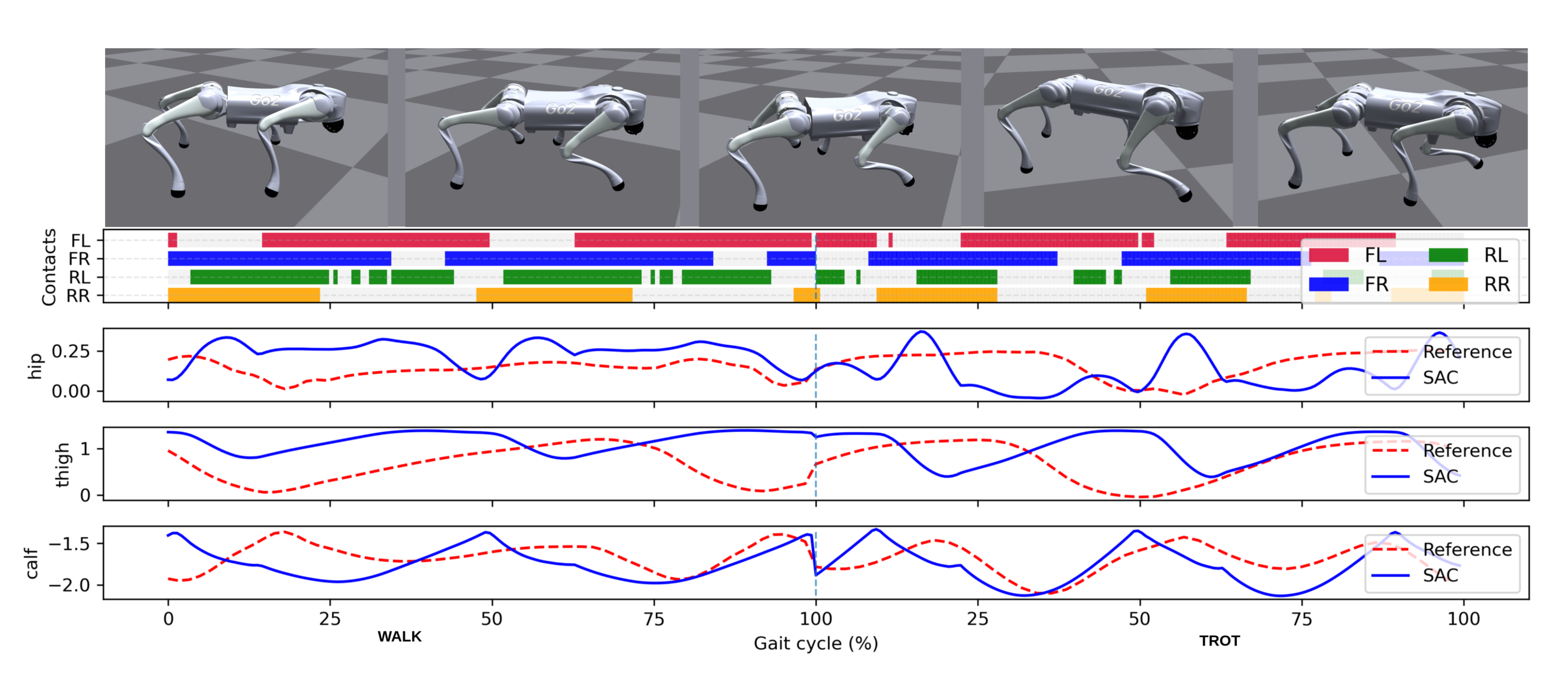}
\caption{Joint rotations and foot contact timings during walk and trot cycles. 
Front-left hip, thigh, and calf joint rotations (rad) over normalized gait cycles (0--100\% walk, 0--100\% trot), 
obtained using the AMP+SAC framework, on flat terrain.}
	\label{fig:Walk_trot_joint_motion}
\end{figure*}
\begin{figure}[h]
	\centering
	\includegraphics[width =0.5\textwidth]{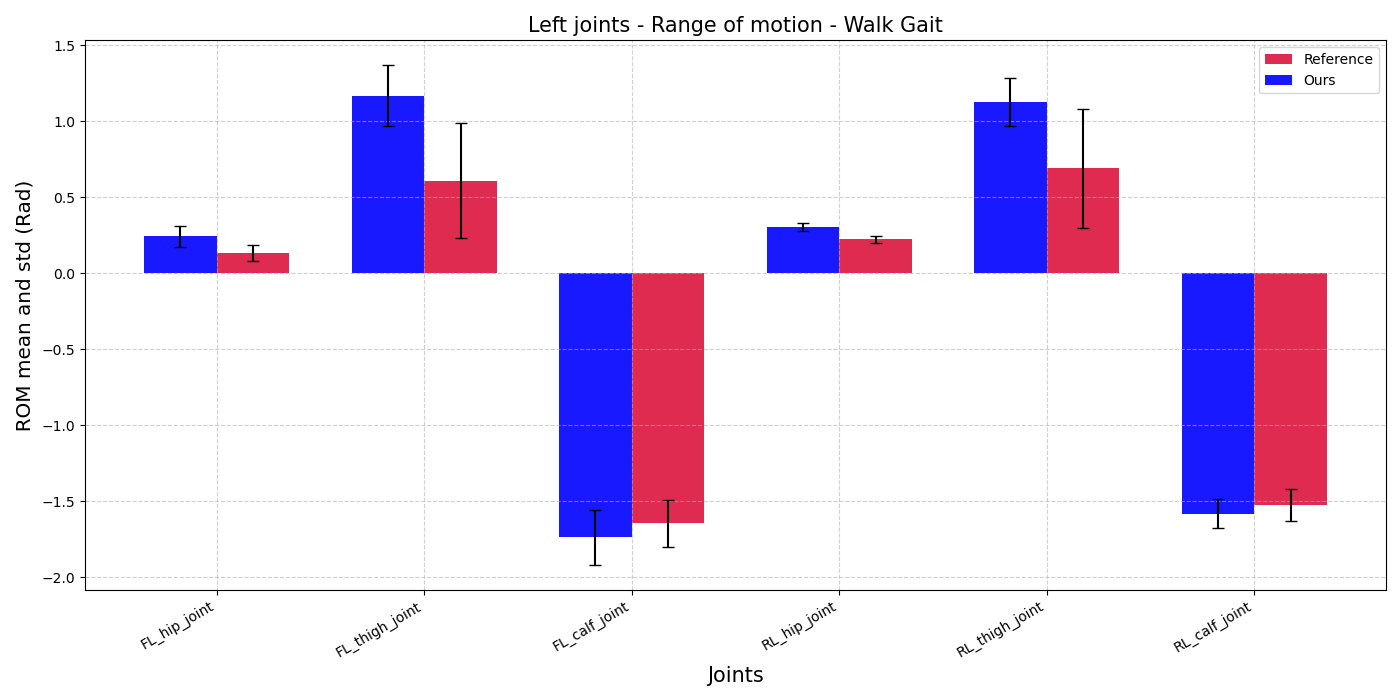}
\caption{Comparison of joint ROM (mean $\pm$ std) for the walk, on the left side, between the reference re-targeted data and our AMP+SAC implementation, on flat terrain.}
	\label{fig:Walk_joint_range_motion_comparisonL}
\end{figure}
\begin{figure}[h]
	\centering
	\includegraphics[width =0.5\textwidth]{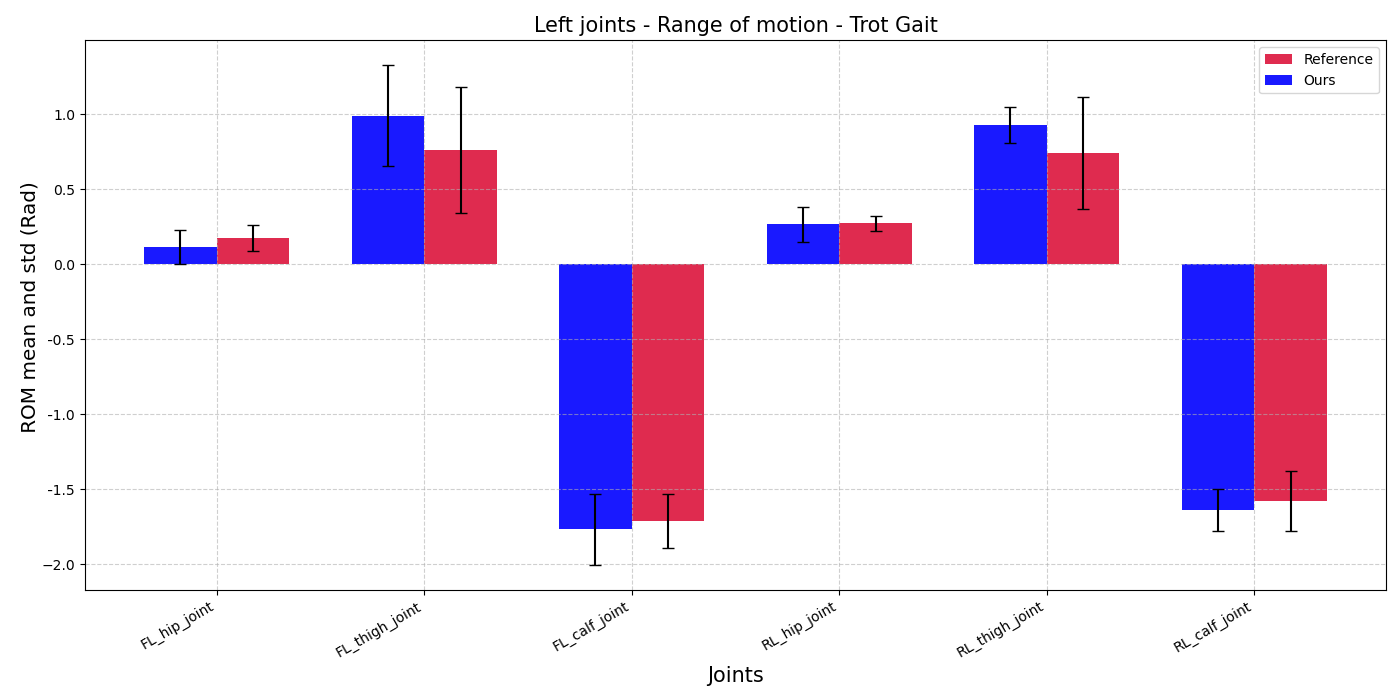}
\caption{Comparison of joint ROM and standard deviation for the trot, on the left side, between the reference re-targeted data and our AMP+SAC implementation, on flat terrain.}
	\label{fig:Trot_joint_range_motion_comparisonL}
\end{figure}
\begin{figure}[h]
	\centering
	\includegraphics[width =0.5\textwidth]{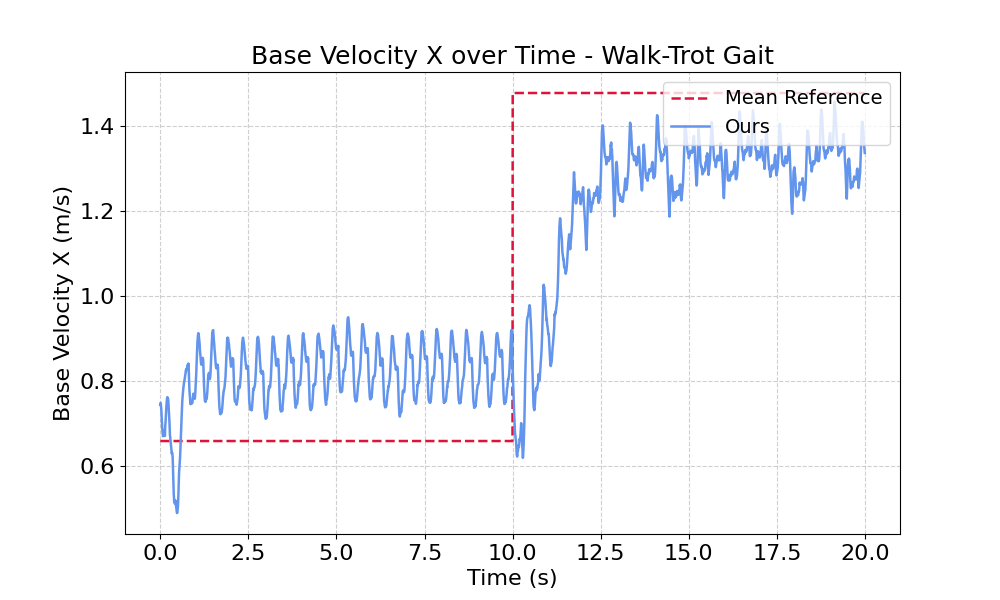}
	\caption{Base velocity in the $x$-direction over time for walk and trot motions, including the transitional velocity obtained with the proposed AMP+SAC implementation. On flat terrain with multiple motions.}
	\label{fig:baseVeloX}
\end{figure}
\begin{figure}[h]
	\centering
	\includegraphics[width =0.5\textwidth]{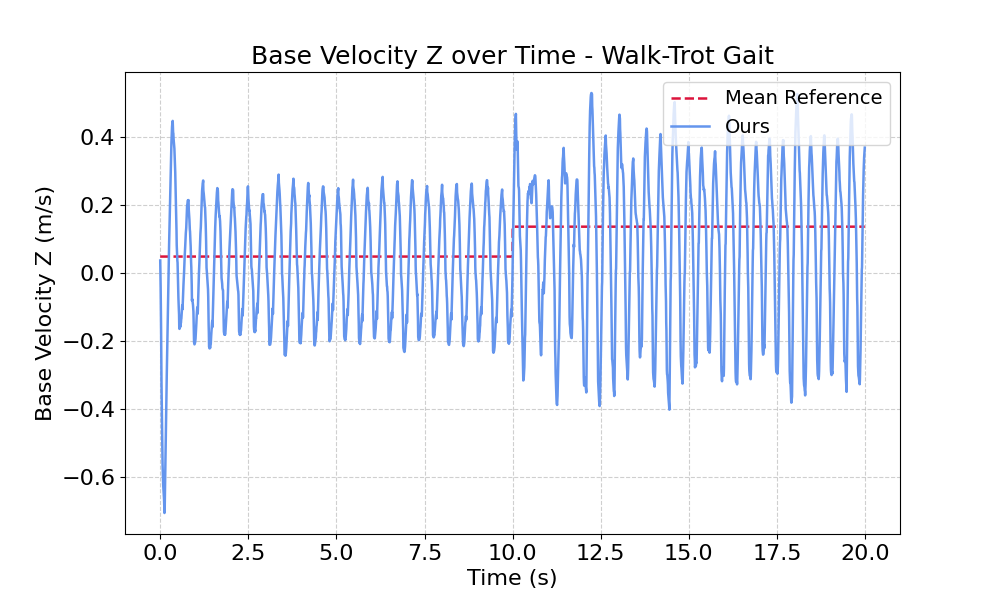}
	\caption{Base velocity in the $z$-direction over time for walk and trot motions, including the transitional velocity obtained with the proposed AMP+SAC implementation. On flat terrain with multiple motions.}
	\label{fig:baseVeloZ}
\end{figure}

\begin{figure}[h]
	\centering
	\includegraphics[width =0.5\textwidth]{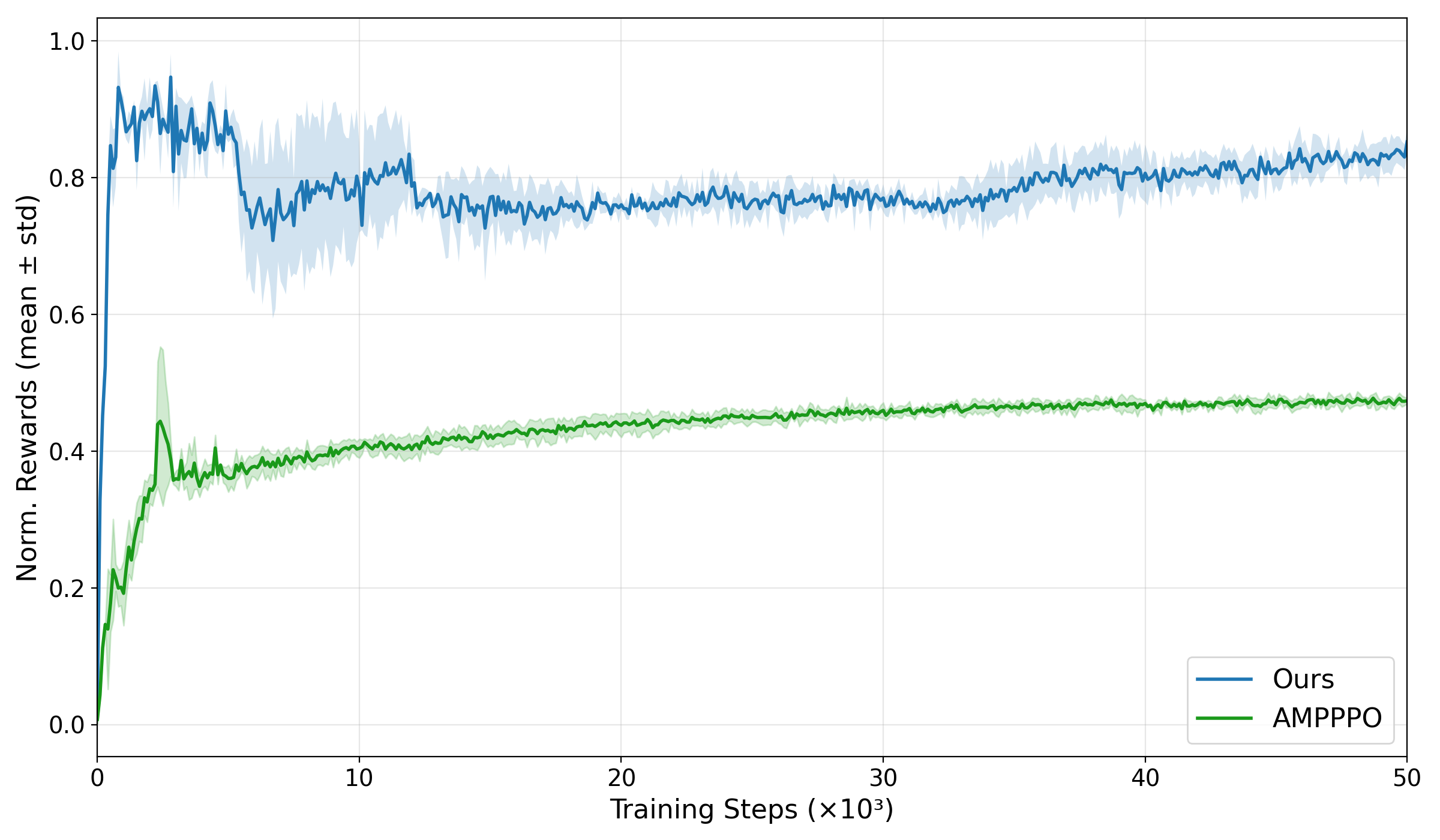}
	\caption{Reward discriminator (mean $\pm$ std) over 50k episodes, AMP+SAC  (ours) vs.\ AMP+PPO. On flat terrain with multiple motions.}
	\label{fig:reward_disc_mean_std_50k_int_norm}
\end{figure}
\begin{figure}[h]
	\centering
	\includegraphics[width =0.5\textwidth]{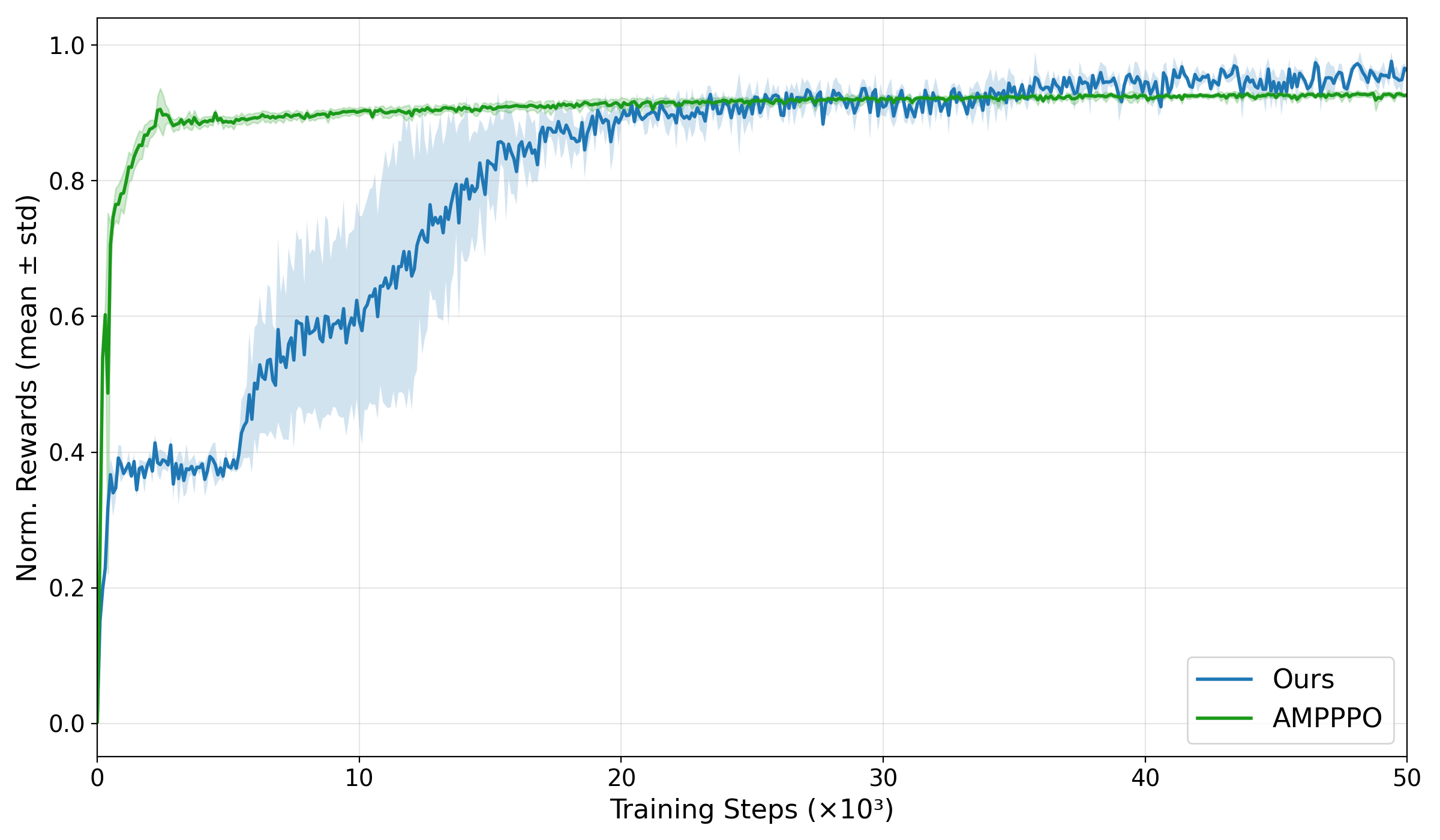}
\caption{Reward (mean $\pm$ std) over 50k episodes, AMP+SAC  (ours) vs.\ AMP+PPO. On flat terrain with multiple motions.}
	\label{fig:reward_mean_std_50k_int_norm}
\end{figure}

\begin{figure}[h]
	\centering
	\includegraphics[width =0.5\textwidth]{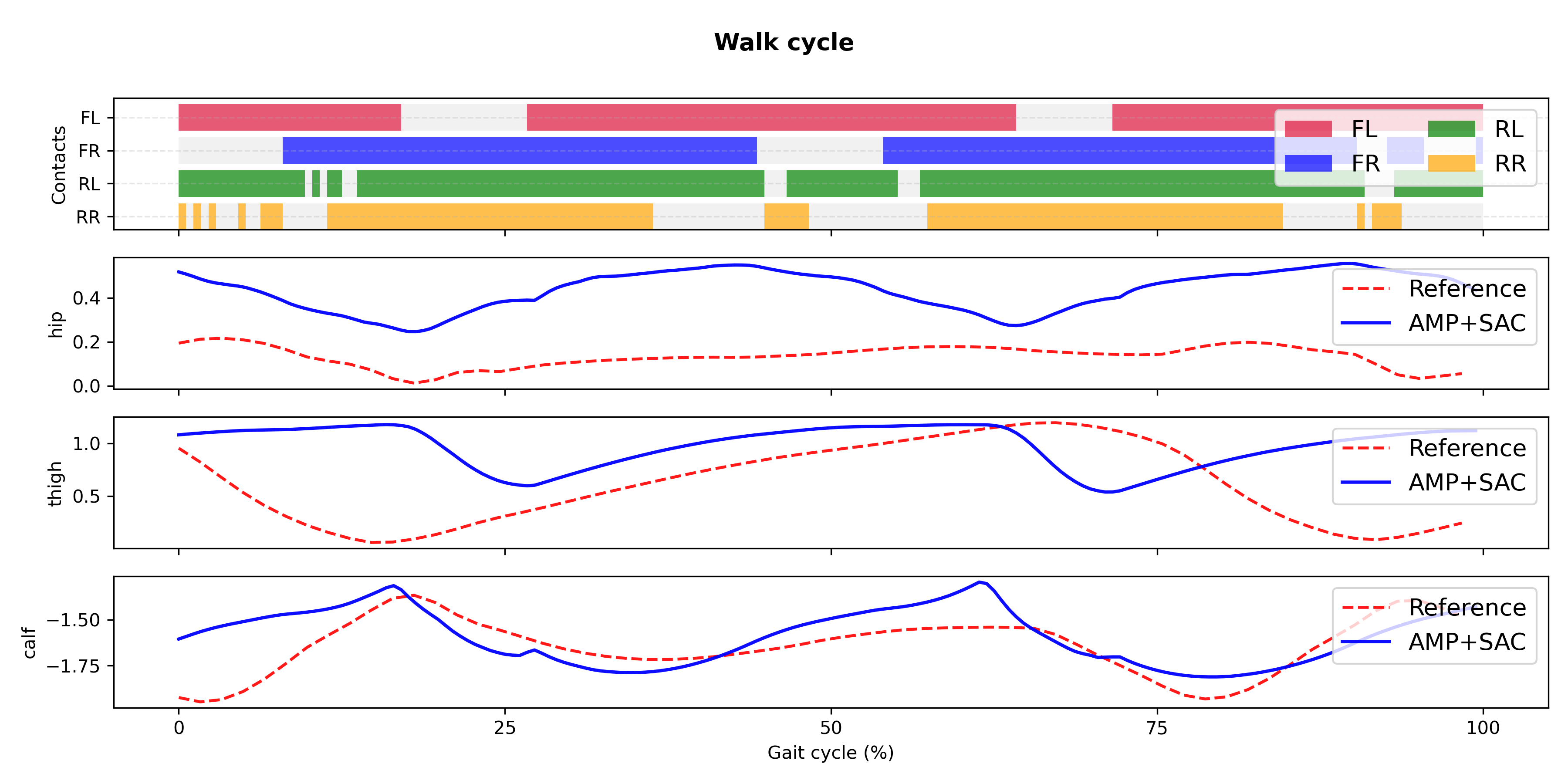}
\caption{Joint rotations and foot contact timings during walk cycle. Front-left hip, thigh, and calf joint rotations (rad) over normalized gait cycles (0--100\% walk), obtained using the proposed AMP+SAC implementation. On wave terrain with a maximum undulation amplitude of 3 cm.}
	\label{fig:Walk_trot_joint_motion_terrain}
\end{figure}
\begin{figure}[h]
	\centering
	\includegraphics[width =0.5\textwidth]{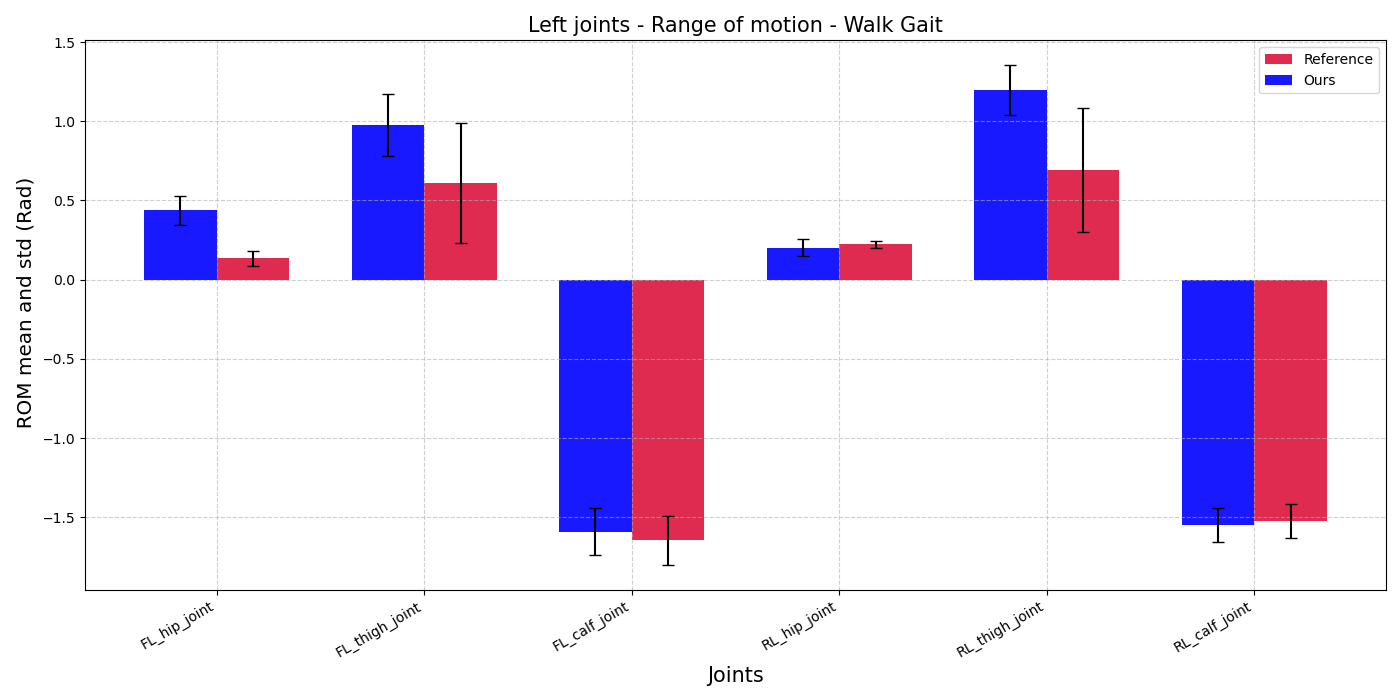}
\caption{Comparison of joints ROM (mean $\pm$ std) over one walk cycle, on the left side, between the reference re-targeted data and our method (AMP+SAC). On wave terrain with a maximum undulation amplitude of 3 cm.}
	\label{fig:Trot_joint_range_motion_comparisonL_terrain}
\end{figure}

\begin{figure}[h]
	\centering
	\includegraphics[width =0.5\textwidth]{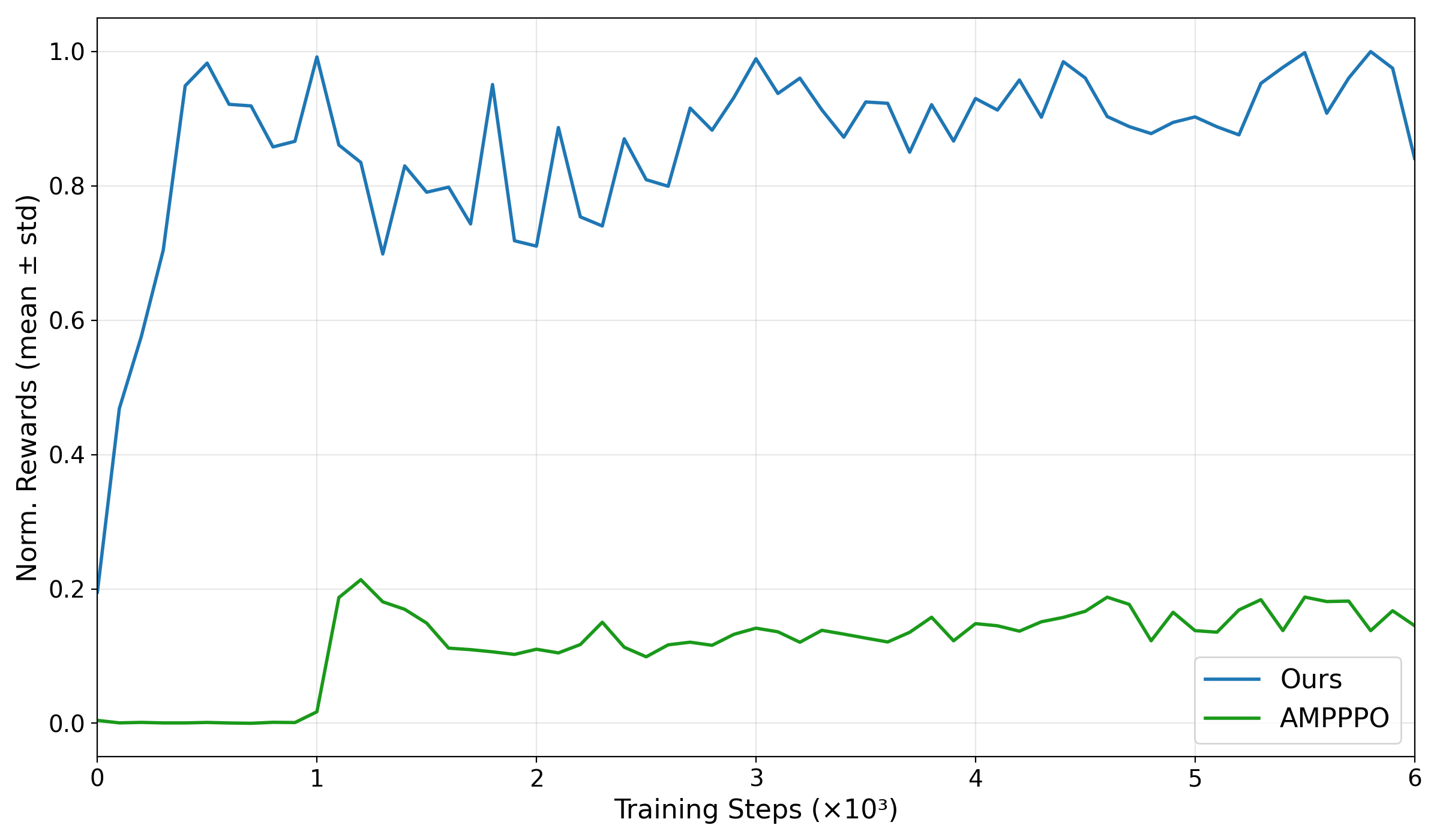}
	\caption{Reward discriminator (mean $\pm$ std) over 6000 episodes, AMP+SAC (ours) vs.\ AMP+PPO. On wave terrain with a maximum undulation amplitude of 3 cm.}
	\label{fig:reward_disc_mean_std_50k_int_norm_terrain}
\end{figure}
\begin{figure}[h]
	\centering
	\includegraphics[width =0.5\textwidth]{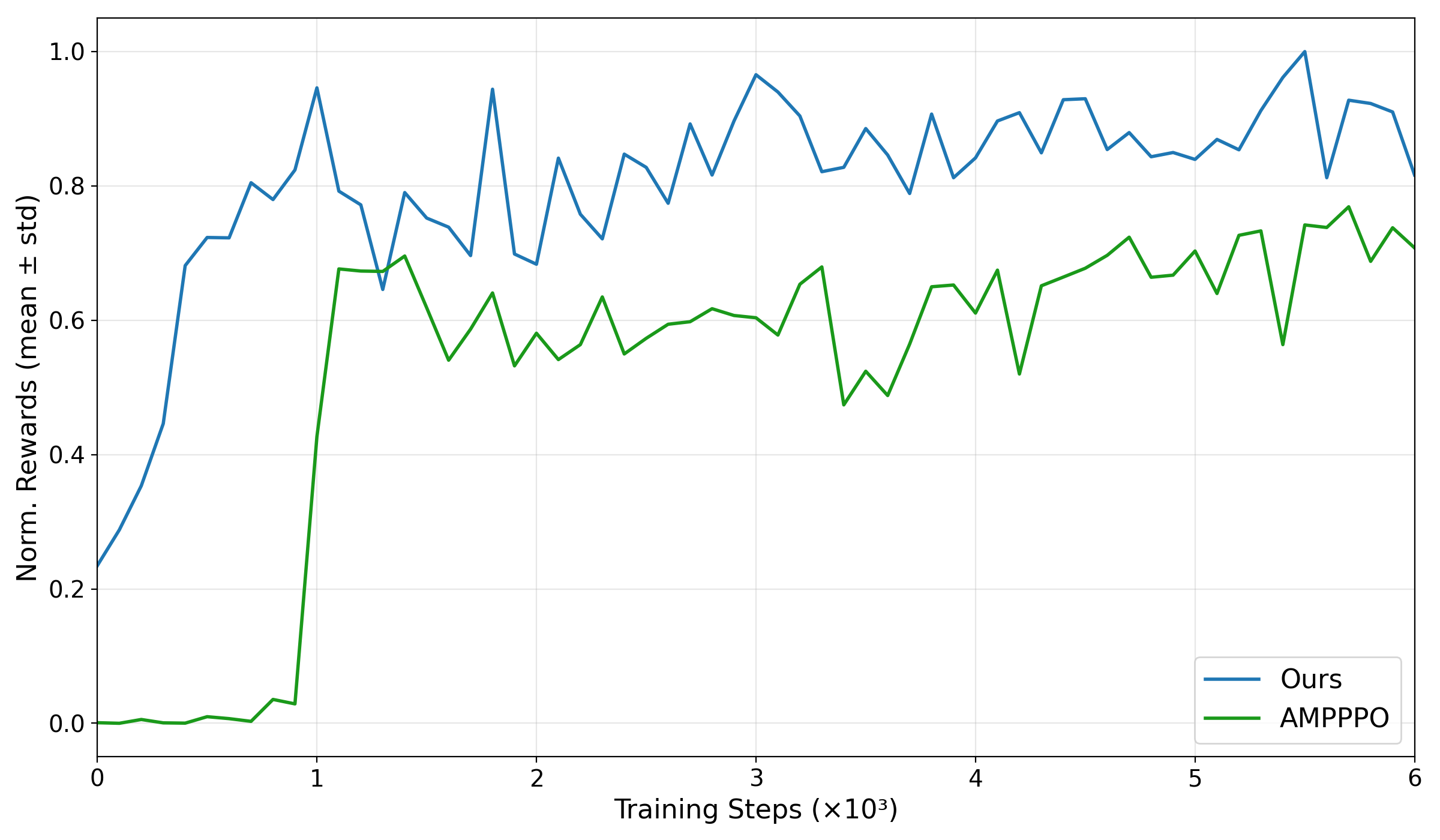}
\caption{Reward (mean $\pm$ std) over 6000 episodes, AMP+SAC (ours) vs.\ AMP+PPO . On wave terrain with a maximum undulation amplitude of 3 cm.}
	\label{fig:reward_mean_std_50k_int_norm_terrain}
\end{figure}	
\section{Experimental implementation}
\label{sec:implementation} 
In this section, we present the experimental setup employed to evaluate the proposed imitation learning framework. 
We detail the training configuration and evaluation protocol used to benchmark SAC and PPO within AMP framework. 
\subsection{Reference Dataset}
The motion capture dataset used in this work \cite{zhang2018mode} contains clips of a German Shepherd dog performing various motion behaviors. 
For this work, we selected walking and trotting gaits, each clipped to one gait cycle. The resulting data have a total duration of 1.95 seconds 
(95 frames), with 1.134 seconds (55 frames) for walking and 0.82 seconds (40 frames) for trotting. 
The raw data consist of 3D joint positions and orientations, which were pre-processed using the methodology detailed in Sec.II.b.
\subsection{Multi-Gait Imitation Learning}
We evaluate the performance of our method against the established benchmark of PPO within the AMP framework. Both algorithms were trained under identical conditions, with 5 random seeds each, using the same reward function parameterization and domain randomization settings. 
Our SAC-based implementation uses separate Actor and Critic networks, each comprising two hidden layers with 1024 and 512 units, and GELU activation functions. The AMP discriminator network also consists of two hidden layers but uses ReLU activation.
We choose AdamW to optimize $\alpha$ and actor. The Actor is optimized with the policy parameters coupled with discriminator's encoder and readout layer,  for alignment. 
The discriminator is optimized within a supervised learning objective defined in Eq. ~\ref{equ:disc}.
Training is performed using a replay buffer with a capacity of $10^6$ experience tuples. For each training iteration, a batch of 16.384 transitions is sampled for the SAC policy update over 8 epochs, while a separate batch of 8.192 transitions is used for a single discriminator update epoch.
Additionally, adopted automatic $\alpha$ tuning to leverage the trade-off between exploration and exploitation based on the entropy level of the policy. The complete hyper-parameter configuration for our method is detailed in Table~\ref{tab:sac-hyperparams} and Table~\ref{tab:amp-hyperparams}. For the PPO+AMP baseline, we adopted the setup and hyper-parameters from \cite{escontrela2022adversarial}. 
\subsection{Progressive Terrain Curriculum}
To evaluate the robustness and generalization capabilities of the learned policies, we employed a structured terrain curriculum. This curriculum progresses from flat ground to continuous wave-like terrains characterized by a maximum undulation height of 3 cm, with varying amplitude and frequency. 
Each policy was warm-started from a pre-trained multi-gait model previously optimized for flat terrain. This initialization strategy accelerates learning on novel terrains by transferring the prior kinematic knowledge encoded in the policy's weights. Specifically, the weights from the pre-trained model were used to initialize the actor network before proceeding with the multi-gait training regimen on the new terrain.
The core training configuration detailed in \tref{tab:sac-hyperparams} and \tref{tab:amp-hyperparams} was maintained from the warm-start model. However, to enhance training stability, we increased the number of discriminator update steps per iteration to four.
The terrain difficulty automatically advanced to the next level once the policy consistently achieved a performance threshold of at least $60\%$ of the maximum possible reward for velocity tracking on the current terrain.
\subsection{Training}
We conducted the imitation learning experiments using 4096 parallel environments per GPU, with policies queried at 33 Hz. All experiments were performed in simulation using Isaac Gym \cite{liang2018gpu}, running on a workstation equipped with an NVIDIA RTX A6000 GPU and 80 GB RAM. The learned policies were evaluated on a Unitree Go2 quadruped robot with 12 actuated joints.
At the beginning of each training episode, the robot is initialized with a random motion sample from the reference dataset $\mathcal{D}$, ensuring diverse starting conditions. The state representation $s \in \mathcal{S}$ includes the command velocity target, joint positions and velocities, and base orientation. The action $a \in \mathcal{A}$ corresponds to continuous joint position targets, which are fed into PD controllers to compute the motor torques. 
For terrain curriculum experiments, the terrain parameters are also included in the observation space.
For all the experiments, the reward composition is defined by weight for the imitation task $\mathcal{W_{AMP}}$ set to $0.6$ and for the reward task set to $0.4$. The detailed locomotion support reward terms and their scales are listed in the  \tref{tab:rewards}. 
In the complex terrain setting, the penalty for approaching joint limits was decreased to $-2$.
Specifically, in the complex terrain experiment, we decreased the penalization to $-2$ to boost exploration and adaptability in the uneven surfaces.
To improve policy robustness, we applied domain randomization over the ranges specified in \tref{tab:domain_rand}, varying the robot’s base mass, the motor gain multipliers of the PD controllers, and the terrain friction coefficients.
The data collection steps for the multi-gait experiment was approximately 6 billion over 40 hours, corresponding to the equivalent of 1.9 years of real-world experience. 
The terrain curriculum experiment involved roughly 6 million steps completed in 48 hours, equivalent to 0.69 years of real-world interaction. Training on terrain curricula proved more computationally demanding, requiring significantly longer simulation time per step.
\begin{table}[h]
\centering
\caption{SAC hyper-parameters used in training.}
\label{tab:sac-hyperparams}
\begin{tabularx}{0.95\linewidth}{l l}
\toprule
\textbf{Parameter} & \textbf{Value} \\
\midrule
Replay memory size & $1.0 \times 10^{7}$ \\
Batch size & $16384$ \\
Updates per step & $8$ \\
$n$-step return & $3$ \\
Target smoothing coefficient $\tau$ & $0.05$ \\
Discount factor $\gamma$ & $0.99$ \\
Actor learning rate $\alpha_{\pi}$ & $0.001$ \\
Entropy coefficient $\alpha$ & automatic \\
Actor hidden layers & [1024, 512] \\
Actor activation & GELU \\
Critic hidden layers & [1024, 512] \\
Critic activation & GELU \\
Warm-up steps & $100$ \\
\bottomrule
\end{tabularx}
\end{table}
\begin{table}[h]
\centering
\caption{AMP hyper-parameters for imitation learning.}
\label{tab:amp-hyperparams}
\begin{tabularx}{0.95\linewidth}{l l}
\toprule
\textbf{Parameter} & \textbf{Value} \\
\midrule
AMP loss coefficient $\lambda_{\text{AMP}}$ & $0.1$ \\
Gradient penalty coefficient $\lambda_{\text{GP}}$ & $0.01$ \\
AMP batch size & $8192$ \\
AMP batch count per update & $1$ \\
AMP reward coefficient & $2.0$ \\
Pre-loaded motion transitions & $2.0 \times 10^{7}$ \\
Discriminator hidden layers & [1024, 512] \\
Minimum normalized std. & [0.01, 0.01, 0.01] \\
\bottomrule
\end{tabularx}
\end{table}
\begin{table}[h!]
\centering
\caption{Reward terms used in the locomotion task.}
\label{tab:rewards}
\begin{tabularx}{0.95\linewidth}{l l}
\toprule
\textbf{Reward term} & \textbf{Value} \\
\midrule
Tracking lin. vel. $x$ & $240.0$ \\
Lin. vel. $z$ & $-15.0$ \\
Angular vel. $xy$ & $-1.0$ \\
Angular vel. $z$ & $-5.0$ \\
Base height & $-2.0$ \\
Collision & $-1.0$ \\
Feet contact forces & $-2.0$ \\
DoF acceleration & $-2.5\times 10^{-7}$ \\
DoF pos. limits & $-4.0$ \\
Torques & $-5.0\times 10^{-5}$ \\
\bottomrule
\end{tabularx}

\end{table}
\begin{table}[h!]
\centering
\caption{Domain randomization parameters used during training.}
\label{tab:domain_rand}
\begin{tabularx}{0.95\linewidth}{l l}
\toprule
\textbf{Parameter} & \textbf{Range} \\
\midrule
Base mass $(m)$ &  [-1.,\, 1.] \\
PD gains factor &  [0.9,\, 1.1] \\
Terrain friction coefficient $(\mu)$ &   [0.25,\, 1.75] \\
\bottomrule
\end{tabularx}
\end{table}
\subsection{Comparison Metrics}
\subsubsection{Imitation metric }
We evaluate our imitation learning performance using two metrics: the \emph{task reward} (total environment reward) 
and the \emph{discriminator reward} from AMP. For a fair comparison between AMP+SAC and AMP+PPO, each seed trajectory is linearly interpolated  onto a common grid every 100 training steps, and then globally normalized. Given $N$ random seeds, the mean and standard deviation at training step $t$ are computed as
\[
\mu_t = \frac{1}{N} \sum_{i=1}^{N} R_{i,t}, 
\qquad
\sigma_t = \sqrt{\frac{1}{N} \sum_{i=1}^{N} \big(R_{i,t} - \mu_t \big)^2},
\]
where $R_{i,t}$ denotes the normalized metric value (task or discriminator reward) 
from seed $i$ at step $t$. This process quantifies both the average performance and its consistency across random initialization.
\subsubsection{Bio-Mechanical Parameters}
We analyze locomotion performance using three representative parameters: foot contacts, which quantify the stance and swing phases of each limb; the base velocity, reflecting forward progression and gait stability; 
and the joint range of motion (ROM) of the front-left hip, which indicates how closely the robot reproduces the kinematic envelope of the reference motion. This multi-faceted approach allows us to evaluate performance across contact dynamics, whole-body trajectory, and detailed kinematic reproduction.
\section{Results}
\label{sec:results}
In this section, we present the experimental results, compare them against the selected baseline, and highlight key insights. 
The reported outcomes are averaged over six independent runs with different random seeds in order to account for variability in training. Only one side of the robot is represented in the figures for brevity.
\subsection{Evaluation across multiple motions}
In this part, we summarize the obtained performance across the tasks considered in our study, using the implementations introduced above.
\fref{fig:Walk_trot_joint_motion} presents the evolution of the front-left hip, thigh, and calf joint rotations along with the corresponding foot contact timings during normalized walk and trot cycles.  The obtained trajectories with the proposed AMP+SAC framework closely track the re-targeted dog's reference motions, both in amplitude and in phase. From the contact patterns, it can be observed that the policy has learned to reproduce the alternating stance–swing transitions that characterize walk and trot, while performing the reward's tasks of walking forward. 
\fref{fig:Walk_joint_range_motion_comparisonL} and \fref{fig:Trot_joint_range_motion_comparisonL} present a comparison of the obtained ROM, including standard deviations (in rad), for the walk and trot motions. These results are evaluated against the corresponding re-targeted ROM derived from dog reference data.
The figures demonstrate that the imitated motions (walk and trot) achieve joint rotations with ROM closely matching those of the reference. This indicates that the robot not only performs the task of walking straight but also effectively reproduces the underlying gait dynamics of the reference motions. Overall, the results validate that the proposed AMP+SAC framework enables the robot to acquire natural and smooth motion patterns across both walk and trot gaits.
\fref{fig:baseVeloX}, \fref{fig:baseVeloZ} illustrates the evolution of the robot’s base velocity in the $x$ and $z$ directions, obtained with the proposed method. The policy regulates the forward velocity consistently across walk and trot motions.
The results indicate that the learned policy is capable of imitating the reference, leading to stable and natural motion. Combined with the joint rotation and foot contact analysis presented above, these findings demonstrate that the proposed framework successfully learns coordinated gait dynamics that closely imitate animal (dog) motion while performing the task of walking forward.
The results in \fref{fig:reward_disc_mean_std_50k_int_norm} and \ref{fig:reward_mean_std_50k_int_norm} indicate that combining SAC with AMP can provide advantages over AMP+PPO. In terms of imitation performance, AMP+SAC achieves higher discriminator rewards on average, suggesting a closer match to expert behaviors. The overall task reward shows a similar tendency, with AMP+SAC generally reaching a bit higher reward mean across runs. These findings suggest that AMP+SAC improves the imitation learning performance for multi-motion tasks compared to AMP+PPO.
\subsection{Evaluation across multi-terrains}
In the second experiments, we assess the robustness of the proposed approach when transitioning from flat to wave terrain. The objective is to determine whether AMP+SAC can perform the task of walking forward, while imitating the reference under terrain perturbations.
Figure~\ref{fig:Walk_trot_joint_motion_terrain} illustrates the joint rotations (in rad) and foot contact timings over one walk cycle, on wave terrain with a maximum undulation amplitude of 3 cm. This figure shows the expected periodic walking pattern and joint trajectories with similar variations as the reference. Figure~\ref{fig:Trot_joint_range_motion_comparisonL_terrain} compares the obtained joints ROM on the left side of the robot with the re-targeted reference, demonstrating that the trajectories remain closely aligned with the reference ROM.
\fref{fig:reward_disc_mean_std_50k_int_norm_terrain} and \fref{fig:reward_mean_std_50k_int_norm_terrain} compare AMP+SAC with AMP+PPO on wave terrain IL. AMP+SAC consistently achieves higher reward discriminator value as well as higher mean task rewards than AMP+PPO. These figures suggest a close alignment of the learned policy with the reference motion and demonstrate that AMP+SAC performs well even under terrain perturbations. Overall, the findings indicate that AMP+SAC provides more stable performance than AMP+PPO for imitation learning, even on uneven terrain. 
\section{Discussion and Outlooks} 
\label{sec:conclusion}
The presented results highlight the use cases achieved in order to assess the usability of the proposed implementation.  
A core strength of SAC is its replay buffer, which retains short- to medium-horizon interaction histories and enables efficient reuse of past experience, leading to more effective exploration and higher sample efficiency. 
Although this off-policy algorithm is a promising RL approach, SAC performance is sensitive to hyperparameters, which has historically constrained its adoption.
In this work, we proposed, to the best of our knowledge, the first implementation of an AMP+SAC imitation learning framework. We tested our implementation in two experiments: (i) learning two motions, and (ii) continuing imitation while the environment, in our case the terrain, is modified.  
The first experiment tested the capacity to learn and imitate two motions with higher mean imitation reward. The comparison was done against the re-targeted motions from a dog and the AMP+PPO \cite{escontrela2022adversarial}.
The second experiment evaluated the robustness and adaptability of our learning policy, 
in an uneven terrain. 
Under these conditions, SAC demonstrated sustained optimization of the expected return across successive epochs, even when presented with novel, unseen observations.
We observed that AMP+SAC require more training time to learn a good policy than its counterpart AMP+PPO. 
The reason is well established, SAC fundamentally is made for exploration. For high-dimensional tasks, this exploratory priority leads to increased sample complexity and extended training periods before policy improvement is observed.
Therefore, the main limitation of our approach is training time, especially compared to PPO. Another limitation is the buffer. The combination of SAC buffer plus AMP buffer requires higher computational resources. This trade-off is decisive for real-world applications.  
The purpose of this work was to propose to the community a new IL implementation, adding more options to our toolbox. The full implementation will be open-sourced and made available for further experiments and optimization of the current implementation.  
At a time when robots are expected to evolve more and more into our everyday lives, systems must become more generalist and less specialized in a single skill or behavior. For this, an off-policy algorithm like SAC is crucial in enabling the learning of multiple skills or motions. Furthermore, the development of new and versatile IL algorithms is essential to produce smoother and more natural motions, thereby allowing robots to interact more effectively with humans in environments designed for human movements and capacities.  
Based on these very promising results, our future work will explore the integration of an additional small storage to bias SAC replay toward recent data for reflects the current policy and changing AMP reward, while keeping the older transitions to avoid forgetting past experience. 
With this change, we aims to speed up learning and stabilize training under non-stationary conditions.
Additionally, we aim to extend our framework capabilities by enabling the multi-skill learning using hierarchical learning in combination with continual learning technique to add new skills without forgetting, and expands its repertory to skills generalization.
Finally, we plan to perform real-world deployment, especially in legged robots, to test the robustness and generalization of our framework addressing sim-to-real challenges.  

%
\bibliographystyle{IEEEtran}
\begin{samepage}
\bibliography{references}
\end{samepage}

\end{document}